\documentclass{article}

\usepackage{times}
\usepackage{graphicx} %
\usepackage{subfigure}
\usepackage{natbib}
\usepackage{algorithm}
\usepackage{algorithmic}
\usepackage{hyperref}
\usepackage{url}
\usepackage{amsmath}
\usepackage{amssymb}
\usepackage{bm}
\usepackage[toc,page]{appendix}
\usepackage{bm,amsthm}

\graphicspath{ {figs/} }

\usepackage{hyperref}

\DeclareMathOperator*{\argmax}{arg\,max}
\DeclareMathOperator*{\argmin}{arg\,min}
\DeclareMathOperator{\aset}{\mathcal{A}}
\DeclareMathOperator{\sset}{\mathcal{S}}

\DeclareMathOperator{\reals}{\mathbb{R}}
\DeclareMathOperator{\action}{\mathbf{a}}
\DeclareMathOperator{\state}{\mathbf{s}}
\DeclareMathOperator{\E}{\mathbb{E}} %
\newtheorem{lemma}{Lemma}

\usepackage[accepted]{icml2015}

\icmltitlerunning{Deep Reinforcement Learning in Large Discrete Action Spaces}

\begin{document} 

\twocolumn[
\icmltitle{Deep Reinforcement Learning in Large Discrete Action Spaces}

\icmlauthor{Gabriel Dulac-Arnold*, Richard Evans*, Hado van Hasselt, Peter Sunehag, Timothy Lillicrap, Jonathan Hunt, Timothy Mann, Theophane Weber, Thomas Degris, Ben Coppin}{dulacarnold@google.com}
\icmladdress{Google DeepMind}

\vskip 0.3in
]

\begin{abstract}

Being able to reason in an environment with a large number of
discrete actions is essential to bringing reinforcement learning to a larger
class of problems.  Recommender systems, industrial plants and language models
are only some of the many real-world tasks involving large numbers of discrete
actions for which current methods are difficult or even often impossible to apply.

An ability to generalize over the set of actions as well as sub-linear
complexity relative to the size of the set are both necessary to handle such
tasks. Current approaches are not able to provide both of these, which motivates
the work in this paper. Our proposed approach leverages prior information about the
actions to embed them in a continuous space upon which it can generalize.
Additionally, approximate nearest-neighbor  methods allow for logarithmic-time
lookup complexity relative to the number of actions, which is necessary for
time-wise tractable training. This combined approach allows reinforcement
learning methods to be applied to large-scale learning problems previously
intractable with current methods.  We demonstrate our algorithm's abilities on a
series of tasks having up to one million actions.

\end{abstract}

\section{Introduction}

Advanced AI systems will likely need to reason with a large number of possible
actions at every step.  Recommender systems used in large systems such as
YouTube and Amazon must reason about hundreds of millions of items every second,
and control systems for large industrial processes may have millions of possible
actions that can be applied at every time step.  All of these systems are
fundamentally reinforcement learning \cite{sutton1998rl} problems, but current
algorithms are difficult or impossible to apply.

In this paper, we present a new policy architecture which operates efficiently
with a large number of actions.  We achieve this by leveraging  prior
information about the actions to embed them in a continuous space upon which the
actor can generalize.  This embedding also allows the policy's complexity to be
decoupled from the cardinality of our action set. Our policy produces a
continuous action within this space, and then uses an approximate nearest-neighbor search to find the set of closest discrete actions in logarithmic time.
We can either apply the closest action in this set directly to the environment, or fine-tune
this selection by selecting the highest valued action in this set relative to a
cost function. This approach allows for generalization over the action set in logarithmic time, which
is necessary for making both learning and acting tractable in time.

We begin by describing our problem space and then detail our policy architecture,
demonstrating how we can train it using policy gradient methods in an actor-critic framework.  We demonstrate the effectiveness of our policy on various
tasks with up to one million actions, but with the intent that our approach
could scale well beyond millions of actions.

\section{Definitions}
\label{sec:definitions}

We consider a Markov Decision Process (MDP) where $\aset$ is the set of discrete
actions, $\sset$ is the set of discrete states, $\mathcal{P} : \sset \times \aset
\times \sset \rightarrow \reals$ is the transition probability distribution, $R
: \sset \times \aset \rightarrow \reals$ is the reward function, and $\gamma \in
[0,1]$ is a discount factor for future rewards.  Each action $\action \in
\aset$ corresponds to an $n$-dimensional vector, such that $\action \in
\reals^n$.  This vector provides information related to the action. In the same
manner, each state $\state \in \sset$ is a vector $\state \in \reals^m$.

The return of an episode in the MDP is the discounted sum of rewards received by
the agent during that episode: $R_t = \sum_{i=t}^T \gamma^{i-t} r(\state_i, \action_i)$. The goal of RL is to learn a policy $\pi : \sset
\rightarrow \aset$ which maximizes the expected return over all episodes,
$\E[R_1]$.  The state-action value function $Q^\pi(\state, \action) = \E[R_1 |
\state_1 = \state, \action_1 = \action, \pi]$ is the expected return starting from a given state
$\state$ and taking an action $\action$, following $\pi$ thereafter.  $Q^\pi$
can be expressed in a recursive manner using the Bellman equation:

\vspace{-0.5cm}
\begin{equation*}
Q^\pi(\state, \action) = r(\state ,\action) + \gamma \sum_{\state'} P(\state'|\state,\action) Q^\pi(\state',\pi(\state')).
\end{equation*}
\vspace{-0.5cm}

In this paper, both $Q$ and $\pi$ are approximated by parametrized functions.

\section{Problem Description}
\label{sec:probdesc}

There are two primary families of policies often used in RL systems: value-based,
 and actor-based policies.

For value-based policies, the policy's decisions are directly conditioned on the
 value function. One of the more common examples is a policy that is
greedy relative to the value function:

\vspace{-0.5cm}
\begin{equation}
\pi_Q(\state) = \argmax_{\action \in \aset} Q(\state,\action).
\label{eq:argmaxPolicy}
\end{equation}
\vspace{-0.5cm}

In the common case that the value function is a parameterized function which
takes both state and action as input, $|\aset|$ evaluations are necessary to
choose an action.  This quickly becomes intractable, especially if the
parameterized function is costly to evaluate, as is the case with deep neural
networks. This approach does, however, have the desirable property of being
capable of generalizing over actions when using a smooth function approximator.  If
$\action_i$ and $\action_j$ are similar, learning about $\action_i$ will also
inform us about $\action_j$.  Not only does this make learning more efficient,
it also allows value-based policies to use the action features to reason about
previously unseen actions.  Unfortunately, execution complexity grows linearly
with $|\aset|$ which renders this approach intractable when the number of
actions grows significantly.

In a standard actor-critic approach, the policy is explicitly defined by a
parameterized actor function: $\pi_\theta : \sset \rightarrow \aset$.  In
practice $\pi_\theta$ is often a classifier-like function approximator, which
scale linearly in relation to the number of actions.  However, actor-based
architectures avoid the computational cost of evaluating a likely costly
Q-function on every action in the $\argmax$ in Equation \eqref{eq:argmaxPolicy}.
Nevertheless, actor-based approaches do not generalize over the action space as
naturally as value-based approaches, and cannot extend to previously unseen
actions.

Sub-linear complexity relative to the action space and an ability to generalize
over actions are both necessary to handle the tasks we interest ourselves with.
Current approaches are not able to provide both of these, which motivates the
approach described in this paper.

\section{Proposed Approach}

\begin{figure}[h]
\begin{center}
\includegraphics{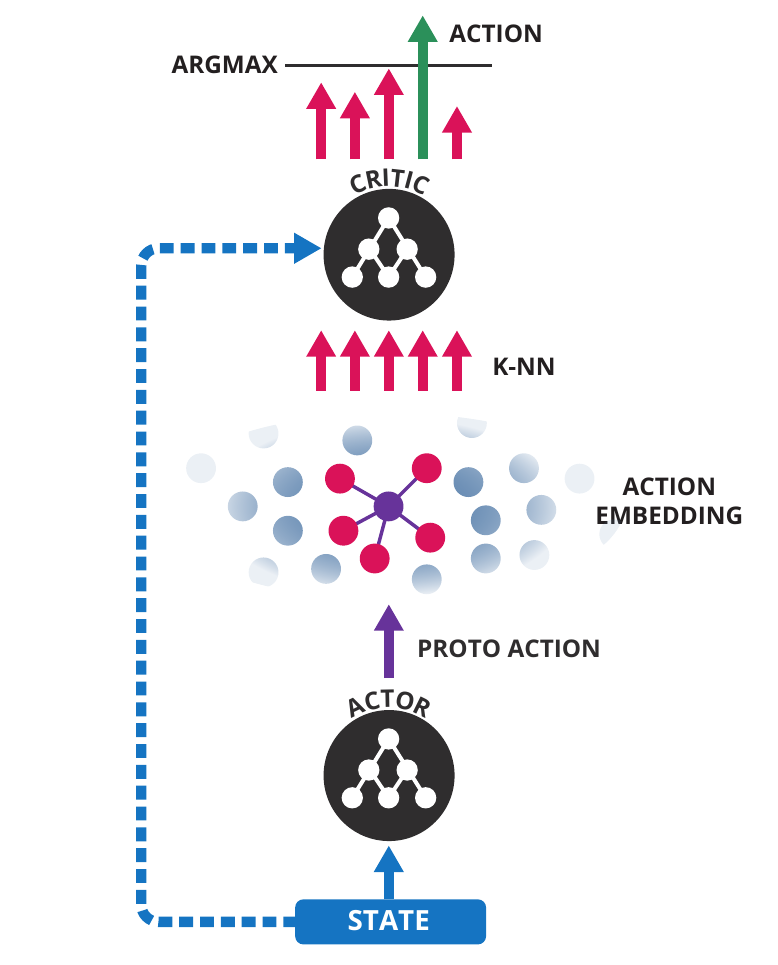}
\end{center}
\caption{Wolpertinger Architecture}
\label{fig:wolpertinger-arch}
\end{figure}

We propose a new policy architecture which we call the
Wolpertinger architecture. This architecture avoids the heavy cost of evaluating
all actions while retaining generalization over actions.  This policy builds upon the actor-critic \cite{sutton1998rl} framework.  We define both an efficient action-generating actor, and
utilize the critic to refine our actor's choices for the full policy. We use multi-layer neural
networks as function approximators for both our actor and critic functions.  We
train this policy using Deep Deterministic Policy Gradient \cite{ddpg}.

The Wolpertinger policy's algorithm is described fully in Algorithm \ref{alg:actor} and
illustrated in Figure \ref{fig:wolpertinger-arch}.  We will detail these in the following sections.

\vspace{-0.25cm}
\begin{algorithm}[h]
  \caption{Wolpertinger Policy}
  \label{alg:actor}
  \begin{algorithmic}
    \STATE State $\state$ previously received from environment.
    \STATE $\hat{\action} = f_{\theta^\pi}(\state)$ \COMMENT{Receive proto-action from actor.}
    \STATE $\aset_k = g_k(\hat{\action})$ \COMMENT{Retrieve $k$ approximately closest actions.}
    \STATE $\action = \argmax_{\action_j \in \aset_k} Q_{\theta^Q}(\state, \action_j)$
    \STATE Apply $\action$ to environment; receive $r, \state'$.
  \end{algorithmic}
\end{algorithm}
\vspace{-0.2cm}

\subsection{Action Generation} 
\label{sec:action-gen}

Our architecture reasons over actions within a continuous space $\reals^n$, and
then maps this output to the discrete action set $\aset$.  We will first define:

\vspace{-0.8cm}
\begin{align*}
&f_{\theta^{\pi}} : \sset \rightarrow \reals^n \\
&f_{\theta^{\pi}}(\state) = \hat{\action}.
\end{align*}
\vspace{-0.8cm}

$f_{\theta^\pi}$ is a function parametrized by $\theta^\pi$, mapping from the state representation space
$\reals^m$ to the action representation space $\reals^n$.  This function
provides a proto-action in $\reals^n$ for a given state, which will likely not
be a valid action, i.e. it is likely that $\hat{\action} \notin \aset$.
Therefore, we need to be able to map from $\hat{\action}$ to an element in
$\aset$.  We can do this with:

\vspace{-0.8cm}
\begin{align*}
&g : \reals^n \rightarrow \aset \\
&g_k(\hat{\action}) = \argmin^k_{\action \in \aset} |\action - \hat{\action}|_2.
\end{align*}
\vspace{-0.4cm}

$g_k$ is a $k$-nearest-neighbor mapping from a continuous space to a discrete set\footnote{For $k=1$ this is a simple nearest neighbor lookup.}. It returns the $k$ actions in $\aset$ that are closest to $\hat{\action}$ by
$L_2$ distance.  In the exact case, this lookup is of the same complexity as
the $\argmax$ in the value-function derived policies described in Section
\ref{sec:probdesc}, but each step of evaluation is an $L_2$ distance instead of
a full value-function evaluation. This task has been extensively studied in the
approximate nearest neighbor literature, and the lookup can be performed in an
approximate manner in logarithmic time \cite{flann_pami_2014}.  This step is
described by the bottom half of Figure \ref{fig:wolpertinger-arch}, where we can
see the actor network producing a proto-action, and the $k$-nearest neighbors
being chosen from the action embedding.

\subsection{Action Refinement}
\label{sec:reranking}

 Depending on how well the action representation is structured, actions with a
low $Q$-value may occasionally sit closest to $\hat{\action}$ even in a part of
the space where most actions have a high $Q$-value. Additionally, certain
actions may be near each other in the action embedding space, but in certain
states they must be distinguished as one has a particularly low long-term value
relative to its neighbors.  In both of these cases, simply selecting the closest
element to $\hat{\action}$ from the set of actions generated previously is not
ideal.

To avoid picking these outlier actions, and to generally improve the finally
emitted action, the second phase of the algorithm, which is described by the top
part of Figure \ref{fig:wolpertinger-arch}, refines the choice of action by
selecting the highest-scoring action according to $Q_{\theta^Q}$:

\vspace{-0.4cm}
\begin{equation}
\label{eq:policy}
\pi_{\theta}(\state) = \argmax_{a \in g_k \circ f_{\theta^\pi} (\state)} Q_{\theta^Q}(\state,\action).
\end{equation}

This equation is described more explicitly in Algorithm \ref{alg:actor}.  It introduces $\pi_\theta$ which is the full Wolpertinger policy. The parameter $\theta$ represents both the parameters of the action generation element in $\theta^\pi$ and of the critic in $\theta^Q$.

As we demonstrate in Section \ref{sec:experiments}, this second pass makes our
algorithm significantly more robust to imperfections in the choice of action
representation, and is essential in making our system learn in certain domains.
The size of the generated action set, $k$, is task specific, and allows for an
explicit trade-off between policy quality and speed.

\subsection{Training with Policy Gradient}

Although the architecture of our policy is not fully differentiable, we argue
that we can nevertheless train our policy by following the policy gradient of
$f_{\theta^\pi}$.  We will first consider the training of a simpler policy, one
defined only as $\tilde{\pi}_{\theta} = g \circ f_{\theta^\pi}$.  In this
initial case we can consider that the policy is $f_{\theta^\pi}$ and that the
effects of $g$ are a deterministic aspect of the environment.  This allows us to
maintain a standard policy gradient approach to train $f_{\theta^\pi}$ on
\textit{its} output $\hat{\action}$, effectively interpreting the effects of $g$
as environmental dynamics.  Similarly, the $\argmax$
operation in Equation \eqref{eq:policy} can be seen as introducing a non-stationary aspect to the environmental dynamics.

\subsection{Wolpertinger Training}

The training algorithm's goal is to find a parameterized policy $\pi_{\theta^*}$
which maximizes its expected return over the episode's length.  To do this, we
find a parametrization $\theta^*$ of our policy which maximizes its expected
return over an episode: $\theta^* = \argmax_\theta \E[R_1 | \pi_\theta]$.

We perform this optimization using Deep Deterministic Policy Gradient (DDPG)
\cite{ddpg} to train both $f_{\theta^\pi}$ and $Q_{\theta^Q}$.  DDPG draws from
two stability-inducing aspects of Deep Q-Networks \cite{mnih15} to extend
Deterministic Policy Gradient \cite{DPG} to neural network function
approximators by introducing a replay buffer \cite{lin1992self} and target
networks.  DPG is similar to work introduced by NFQCA \cite{NFQCA} and leverages
the gradient-update originally introduced by ADHDP \cite{prokhorov1997adaptive}.

The goal of these algorithms is to perform policy iteration by alternatively
performing policy evaluation on the current policy with Q-learning, and then
improving upon the current policy by following the policy gradient.

The critic is trained from samples stored in a replay buffer \cite{mnih15}.
Actions stored in the replay buffer are generated by $\pi_{\theta^\pi}$, but the
policy gradient $\nabla_{\action} Q_{\theta^Q}(\state, \action)$ is taken at
$\hat{\action} = f_{\theta^\pi}(\state)$.  This allows the learning algorithm to
leverage the otherwise ignored information of which action was actually executed for
training the critic, while taking the policy gradient at the actual output of
$f_{\theta^\pi}$. The target action in the Q-update is generated by the full policy $\pi_\theta$ and not simply $f_{\theta^\pi}$.

A detailed description of the algorithm is available in the supplementary material.

\section{Analysis}
\label{sec:dim_returns}
Time-complexity of the above algorithm scales linearly in the number of selected actions, $k$.  We will see that in practice though, increasing $k$ beyond a certain limit does not provide increased performance.  There is a diminishing returns aspect to our approach that provides significant performance gains for the initial increases in $k$, but quickly renders additional performance gains marginal.

Consider the following simplified scenario.  For a random proto-action $\hat{a}$, each nearby action has a probability $p$ of being a bad or broken action with a low value of $Q(s,\hat{a}) - c$.  The values of the remaining actions are uniformly drawn from the interval $[Q(s,\hat{a})-b, Q(s,\hat{a})+b]$, where $b\le c$.  The probability distribution for the value of a chosen action is therefore the mixture of these two distributions.
\begin{lemma}
\label{eq:lemma}
Denote the closest $k$ actions as integers $\{1, \ldots, k\}$.  Then in the scenario as described above, the expected value of the maximum of the $k$ closest actions is
\begin{align*}
& \mathbb{E}\left[\max_{i\in \{1, \ldots k\}} Q(s,i) \mid s, \hat{a} \right] = Q(s,a) + b  \\
& \qquad - p^k (c - b) - \frac{2b}{k+1} \frac{1 - p^{k+1}}{ 1 - p } 
\end{align*}
\end{lemma}

The highest value an action can have is $Q(s,\hat{a}) + b$. The best action within the $k$-sized set is thus, in expectation, $p^k (c - b) + \frac{2 b}{k+1} \frac{1 - p^{k+1}}{ 1 - p }$ smaller than this value.

The first term is in $O(p^k)$ and decreases exponentially with $k$. The second term is in $O(\frac{1}{k+1})$.   Both terms decrease a relatively large amount for each additional action while $k$ is small, but the marginal returns quickly diminish as $k$ grows larger.  This property is also observable in experiments in Section \ref{sec:experiments}, notably in Figures \ref{fig:835} \& \ref{fig:13k-nn}.  Using $5\%$ or $10\%$ of the maximal number of actions the performance is similar to when the full action set is used.  Using the remaining actions would result in relatively small performance benefits while increasing computational time by an order of magnitude.

The proof to Lemma \ref{eq:lemma} is available in the supplementary material.

\section{Related Work}

There has been limited attention in the literature with regards to large
discrete action spaces within RL.  Most prior work has been concentrated on
factorizing the action space into binary subspaces.  Generalized value functions
were proposed in the form of H-value functions \cite{pazis2011generalized},
which allow for a policy to evaluate $\log(|\aset|)$ binary decisions to act.  This learns a factorized value function
from which a greedy policy can be derived for each subspace.  This
amounts to performing $\log(|\aset|)$ binary operations on each action-selection step.

A similar approach was proposed which leverages Error-Correcting Output Code classifiers
(ECOCs) \cite{dietterich1995solving} to factorize the policy's action space and
allow for parallel training of a sub-policy for each action sub-space
\cite{dulac2012fast} .  In the ECOC-based approach case, a policy is learned
through Rollouts Classification Policy Iteration
\cite{lagoudakis2003reinforcement}, and the policy is defined as a multi-class
ECOC classifier. Thus, the policy directly predicts
a binary action code, and then a nearest-neighbor lookup is performed according
to Hamming distance.

Both these approaches effectively factorize the action space into $\log(|\aset|)$
binary subspaces, and then reason about these subspaces independently.
These approaches can scale to very large action spaces, however, they require a
binary code representation of each action, which is difficult to design
properly. Additionally, the generalized value-function approach uses a Linear
Program and explicitly stores the value function per state, which prevents it
from generalizing over a continuous state space.  The ECOC-based approach only
defines an action producing policy and does not allow for refinement with a $Q$-function.

These approaches cannot naturally deal with discrete actions that have
associated continuous representations.  The closest approach in the literature
uses a continuous-action policy gradient method to learn a policy in a continuous
action space, and then apply the nearest discrete action \cite{van2009using}.
This is in principle similar to our approach, but was only tested on small
problems with a uni-dimensional continuous action space (at most 21 discrete
actions) and a low-dimensional observation space.  In such small discrete action
spaces, selecting the nearest discrete action may be sufficient, but we show in
Section \ref{sec:experiments} that a more complex action-selection scheme is
necessary to scale to larger domains.

Recent work extends Deep Q-Networks to `unbounded' action spaces
\cite{he2015deep}, effectively generating action representations for any action
the environment provides, and picking the action that provides the highest Q.
However, in this setup, the environment only ever provides a small (2-4) number
of actions that need to be evaluated, hence they do not have to explicitly pick
an action from a large set.

This policy architecture has also been leveraged by the authors for learning to
attend to actions in MDPs which take in multiple actions at each state (Slate
MDPs) \cite{sunehag2015deep}.

\section{Experiments}
\label{sec:experiments}
We evaluate the Wolpertinger agent on three environment classes: Discretized Continuous Control, Multi-Step Planning, and Recommender Systems.  These are outlined below:

\subsection{Discretized Continuous Environments}

To evaluate how the agent's performance and learning speed relate to the number
of discrete actions we use the MuJoCo \cite{todorov2012mujoco} physics
simulator to simulate the classic continuous control tasks cart-pole \cite{}.  Each
dimension $d$ in the original continuous control action space is discretized
into $i$ equally spaced values, yielding a discrete action space with $|\aset| =
i^d$ actions.

In cart-pole swing-up, the agent must balance a pole attached to a cart by
applying force to the cart. The pole and cart start in a random downward position, and a
reward of +1 is received if the pole is within 5 degrees of vertical and the
cart is in the middle 10\% of the track, otherwise a reward of zero is received.
The current state is the position and velocity of the cart and pole as well as
the length of the pole.  The environment is reset after 500 steps.

We use this environment as a demonstration both that our agent is able to reason
with both a small and large number of actions efficiently, especially when the
action representation is well-formed.  In these tasks, actions are represented
by the force to be applied on each dimension.  In the cart-pole case, this is
along a single dimension, so actions are represented by a single number. 

\subsection{Multi-Step Plan Environment}

Choosing amongst all possible $n$-step plans is a general large action problem.
For example, if an environment has $i$ actions available at each time step and
an agent needs to plan $n$ time steps into the future then the number of actions
$i^n$ is quickly intractable for $\argmax$-based approaches.  We implement a
version of this task on a puddle world environment, which is a grid world with
four cell types: empty, puddle, start or goal.  The agent consistently starts in
the start square, and a reward of -1 is given for visiting an empty square, a
reward of -3 is given for visiting a puddle square, and a reward of 250 is given
and the episode ends if on a goal cell. The agent observes a fixed-size square
window surrounding its current position.

The goal of the agent is to find the shortest path to the goal that trades off
the cost of puddles with distance traveled.  The goal is always placed in the
bottom right hand corner of the environment and the base actions are restricted
to moving right or down to guarantee goal discovery with random exploration. The
action set is the set of all possible $n$-length action sequences.  We have 2
base actions: $\{\text{down}, \text{right}\}$.  This means that environments
with a plan of length $n$ have $2^n$ actions in total, for $n=20$ we have
$2^{20} \approx 1e6$ actions.

This environment demonstrates our agent's abilities with very large number of
actions that are more difficult to discern from their representation, and have less
obvious continuity with regards to their effect on the environment compared to
the MuJoCo tasks.  We represent each action with the concatenation of each step
of the plan.  There are two possible steps which we represent as either
$\{0,1\}$ or $\{1,0\}$.  This means that a full plan will be a vector of
concatenated steps, with a total length of $2n$.  This representation was chosen
 arbitrarily, but we show that our algorithm is nevertheless able to reason
well with it.

\subsection{Recommender Environment}

To demonstrate how the agent would perform on a real-world large action space
problem we constructed a simulated recommendation system utilizing data from a
live large-scale recommendation engine.  This environment is characterized by a
set of items to recommend, which correspond to the action set $\aset$ and a
transition probability matrix $W$, such that $W_{i,j}$ defines the probability
that a user will accept recommendation $j$ given that the last item they
accepted was item $i$.  Each item also has a reward $r$ associated with it if
accepted by the user.  The current state is the item the user is currently
consuming, and the previously recommended items do not affect the current
transition.

At each time-step, the agent presents an item $i$ to the user with action $\aset_i$.
The recommended item is then either accepted by the user (according to the
transition probability matrix) or the user selects a random item instead. If the
presented item is accepted then the episode ends with probability 0.1, if the
item is not accepted then the episode ends with probability 0.2.  This has the
effect of simulating user patience - the user is more likely to finish their
session if they have to search for an item rather than selecting a
recommendation. After each episode the environment is reset by selecting a
random item as the initial environment state.

\subsection{Evaluation}

 For each environment, we vary the number of nearest neighbors $k$ from $k=1$,
which effectively ignores the re-ranking step described in Section
\ref{sec:reranking}, to $k=|\aset|$, which effectively ignores the action
generation step described in Section \ref{sec:action-gen}.  For $k=1$, we
demonstrate the performance of the nearest-neighbor element of our policy $g \circ
f_{\theta^\pi}$.  This is the fastest policy configuration, but as we see in
the section, is not always sufficiently expressive.  For $k=|\aset|$, we
demonstrate the performance of a policy that is greedy relative to $Q$, always
choosing the true maximizing action from $\aset$.  This gives us an upper bound
on performance, but we will soon see that this approach is often computationally
intractable. Intermediate values of $k$ are evaluated to demonstrate the
performance gains of partial re-ranking.

We also evaluate the performance in terms of training time and average reward
for full nearest-neighbor search, and three approximate nearest neighbor
configurations. We use FLANN \cite{flann_pami_2014} with three settings we refer
to as `Slow', `Medium' and `Fast'.  `Slow' uses a hierarchical k-means tree with
a branching factor of 16, which corresponds to 99\% retrieval accuracy on the
recommender task. `Medium' corresponds to a randomized K-d tree where 39 nearest
neighbors at the leaf nodes are checked. This corresponds to a 90\% retrieval
accuracy in the recommender task. `Fast' corresponds to a randomized K-d tree
with 1 nearest neighbor at the leaf node checked.  This corresponds to a 70\%
retrieval accuracy in the recommender task.  These settings were obtained with
FLANN's auto-tune mechanism.

\section{Results}

In this section we analyze results from our experiments with the
environments described above.  

\subsection{Cart-Pole}

The cart-pole task was generated with a discretization of one million
actions. On this task, our algorithm is able to find optimal policies. We have a
video available of our final policy with one million actions, $k=1$, and `fast'
FLANN lookup here: \url{http://goo.gl/3YFyAE}.

\begin{figure}[h]
\includegraphics[width=8cm]{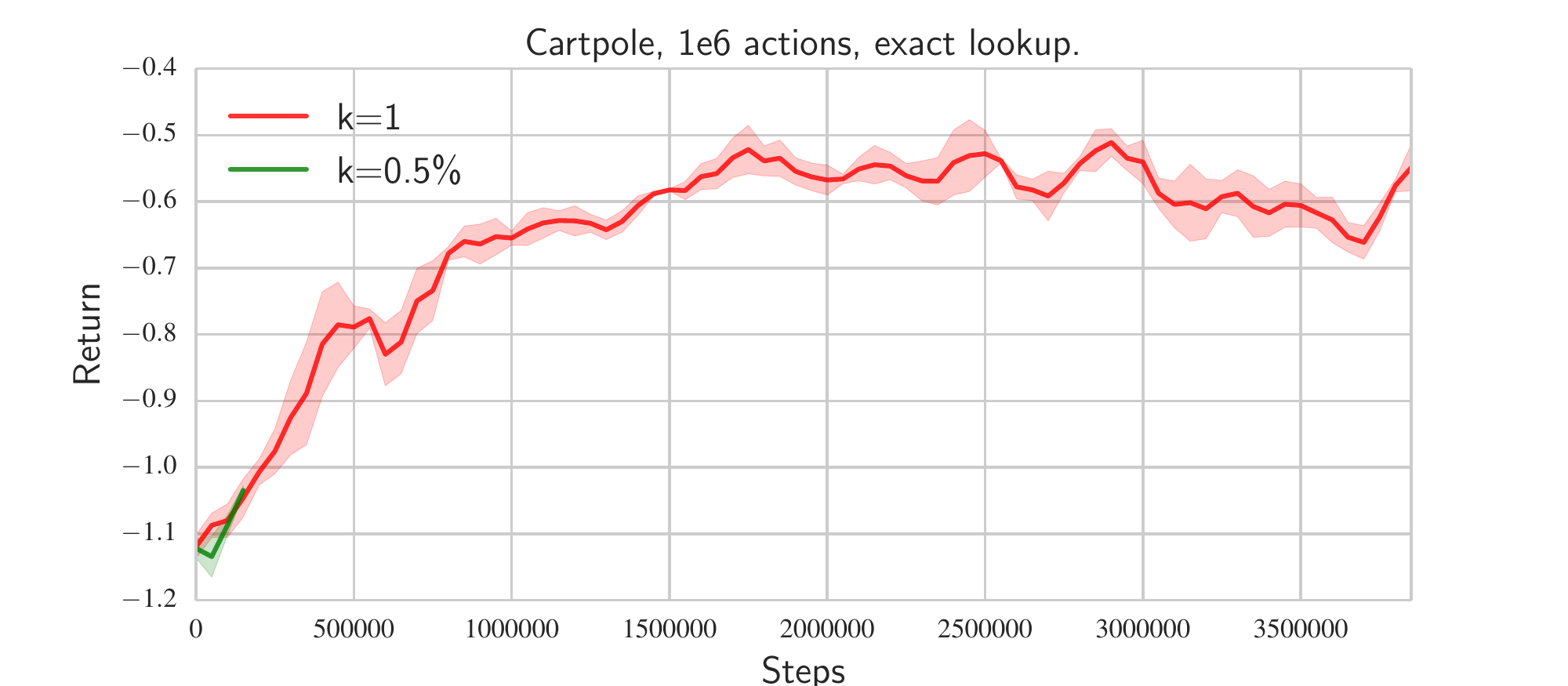}

\caption{Agent performance for various settings of $k$ with exact lookup as a function of steps.  With 0.5\% of neighbors, training time is prohibitively slow and convergence is not achieved.} 
\label{fig:cartpole1}
\end{figure}

We visualize performance of our agent on a one million action cart-pole task with $k=1$ and $k=0.5\%$ in Figure \ref{fig:cartpole1}, using an exact lookup.  In the relatively simple cart-pole task the $k=1$ agent is able to converge to a good policy.  However, for $k=0.5\%$, which equates to 5,000 actions, training has failed to attain more than 100,000 steps in the same amount of time.

\begin{figure}[h]
\includegraphics[width=8cm]{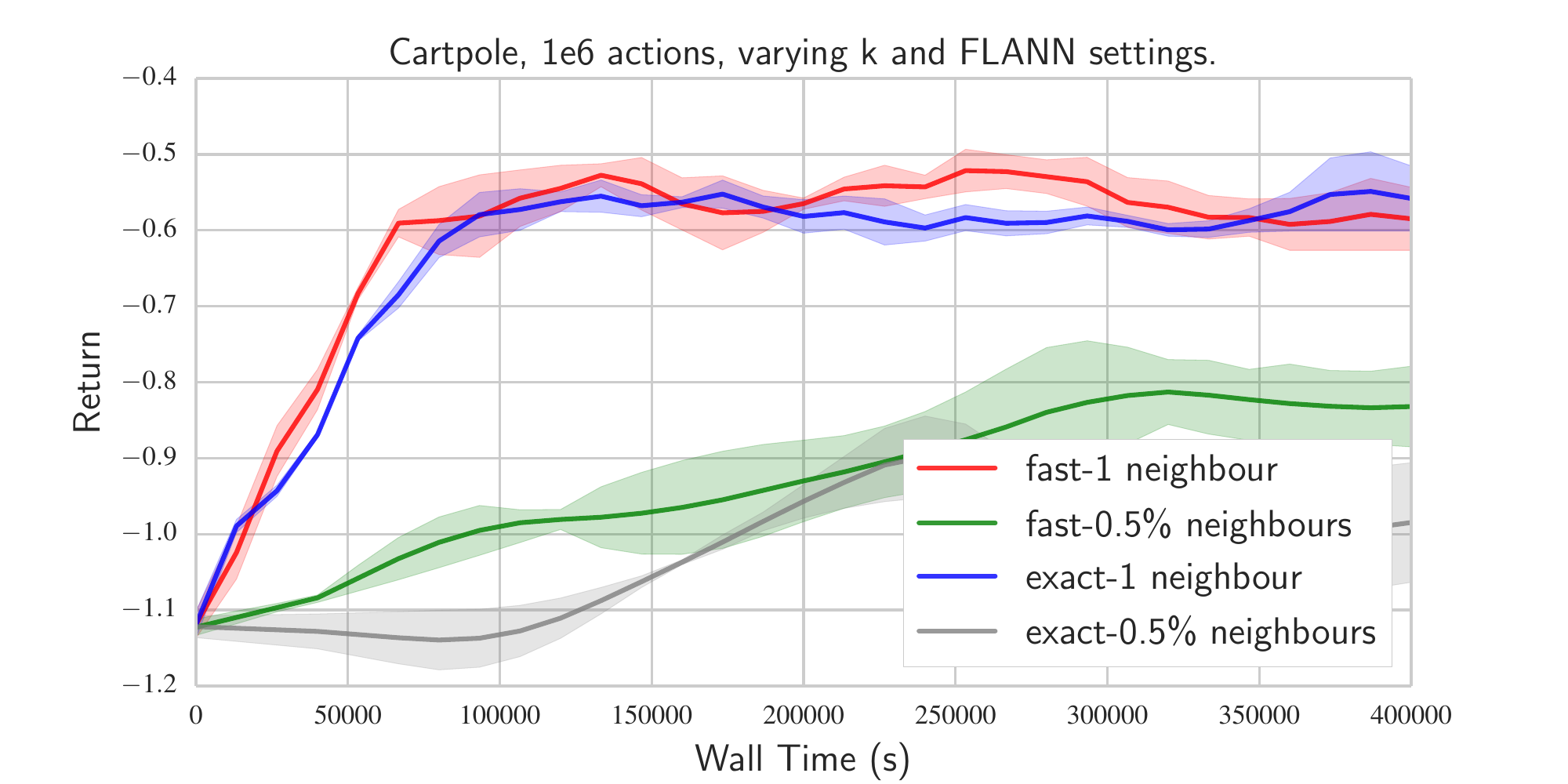}
\caption{Agent performance for various settings of $k$ and FLANN as a function of wall-time on one million action cart-pole.  We can see that with 0.5\% of neighbors, training time is prohibitively slow.} 
\label{fig:cartpole}
\end{figure}

Figure \ref{fig:cartpole} shows performance as a function of wall-time on the
cart-pole task.  It presents the performance of agents with varying neighbor
sizes and FLANN settings after the same number of seconds of training.  Agents
with $k=1$ are able to achieve convergence after 150,000 seconds whereas
$k=5,000$ (0.5\% of actions) trains much more slowly.

\begin{table}[h]
\centering
\begin{tabular}{r||r|r|r|r}
\# Neighbors & Exact & Slow & Medium & Fast   \\
\hline
1 & 18  & 2.4 & 8.5 & 23   \\
0.5\% -- $5,000$ & 0.6 & 0.6 & 0.7 & 0.7  \\
\end{tabular}
\caption{Median steps/second as a function of $k$ \& FLANN settings on cart-pole.}
\label{table:cart}
\end{table}

Table \ref{table:cart} display the median steps per second for the training
algorithm. We can see that FLANN is only helpful for $k=1$ lookups.  Once
$k=5,000$, all the computation time is spent on evaluating $Q$ instead of
finding nearest neighbors.  FLANN performance impacts nearest-neighbor lookup
negatively for all settings except `fast' as we are looking for a nearest
neighbor in a single dimension.  We will see in the next section that for more
action dimensions this is no longer true.

\subsection{Puddle World}

\begin{figure}[h]
\includegraphics[width=8cm]{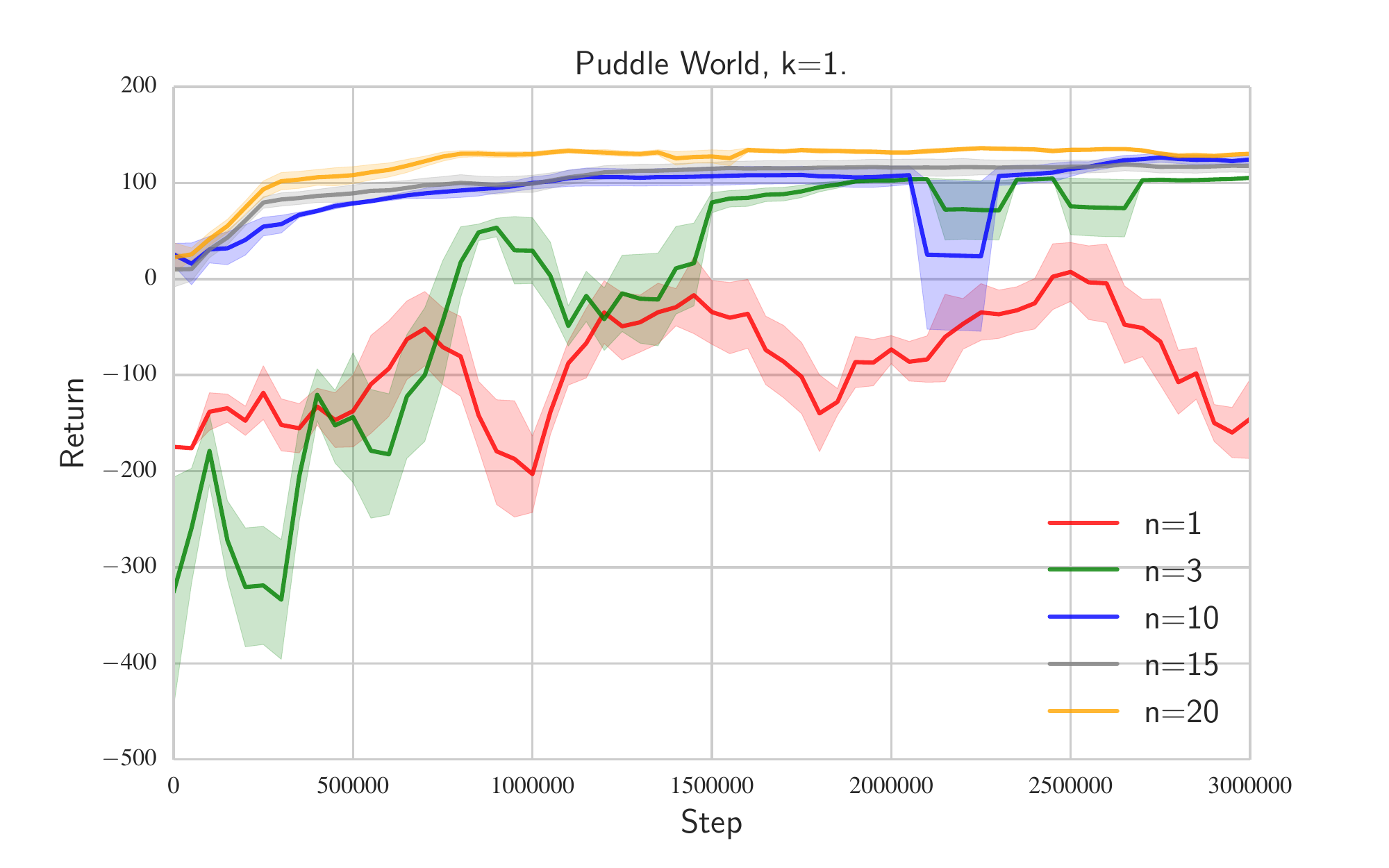}
\caption{Agent performance for various lengths of plan, a plan of $n=20$ corresponds to $2^{20}=1,048,576$ actions.  The agent is able to learn faster with longer plan lengths. $k=1$ and `slow' FLANN settings are used.} 
\label{fig:puddle1}
\end{figure}

We ran our system on a fixed Puddle World map of size $50 \times 50$.  In our
setup the system dynamics are deterministic, our main goal being to show that
our agent is able to find appropriate actions amongst a very large set (up to
more than one million).  To begin with we note that in the simple case with two
actions, $n=1$ in Figure \eqref{fig:puddle1} it is difficult to find a
stable policy.  We believe that this is due to a large number of states
producing the same observation, which makes a high-frequency policy more
difficult to learn.  As the plans get longer, the policies get significantly
better. The best possible score, without puddles, is 150 (50+50 steps of -1,
and a final score of 250).

\begin{figure}[h]
\includegraphics[width=8cm]{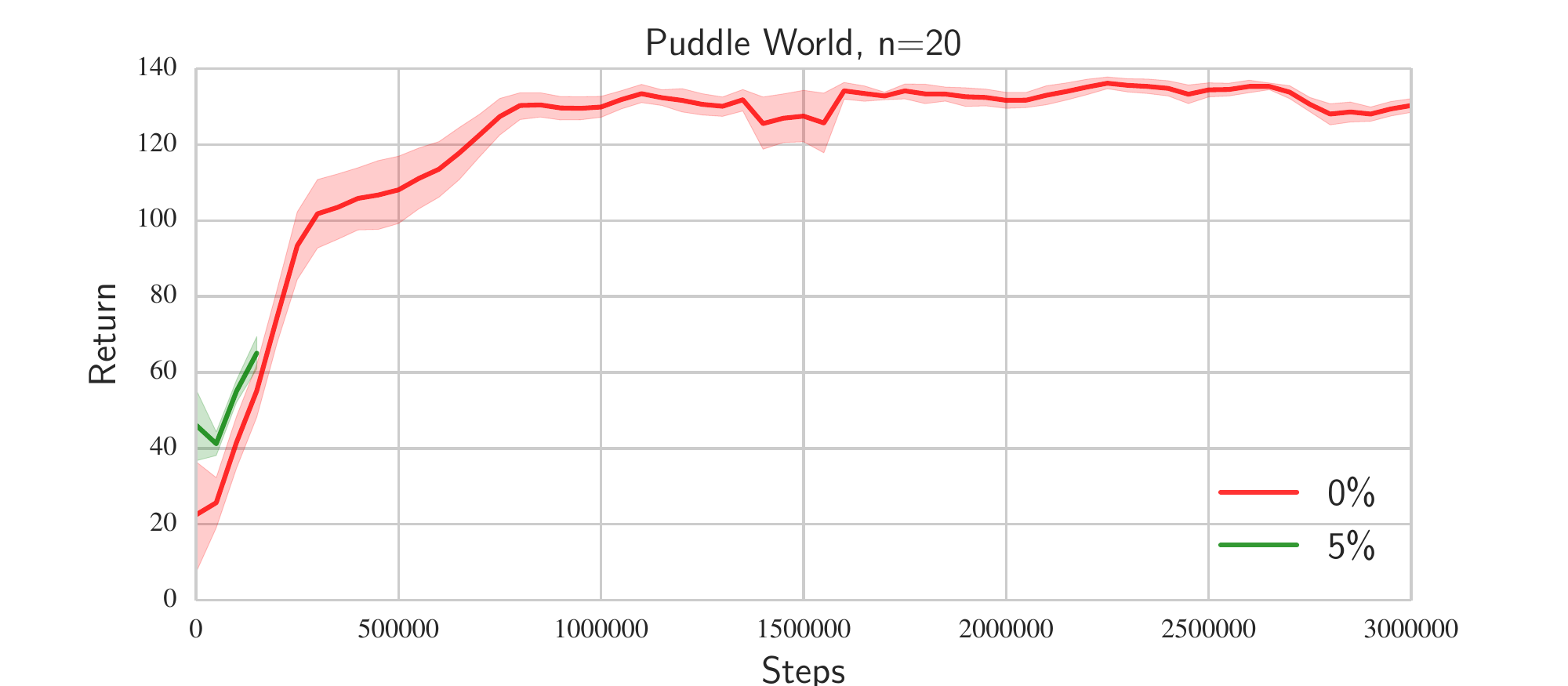}
\caption{Agent performance for various percentages of $k$ in a 20-step plan task in Puddle World with FLANN settings on `slow'.}
\label{fig:puddle20-k}
\end{figure}

Figure \eqref{fig:puddle20-k} demonstrates performance on a 20-step plan Puddle
World with the number of neighbors $k=1$ and $k=52428$, or $5\%$ of actions. In
this figure $k=|\aset|$ is absent as it failed to arrive to the first evaluation
step. We can see that in this task we are finding a near optimal policy while
never using the $\argmax$ pass of the policy. We see that even our most lossy
FLANN setting with no re-ranking converges to an optimal policy in this task.
As a large number of actions are equivalent in value,  it is not surprising that
even a very lossy approximate nearest neighbor search returns sufficiently
pertinent actions for the task.  Experiments on the recommender system in
Section \ref{sec:exp-rec} show that this is not always this case.

Table \ref{table:puddle} describes the median steps per second during training.
In the case of Puddle World, we can see that we can get a speedup for equivalent
performance of up to 1,250 times.

\begin{table}[h]
\centering
\begin{tabular}{r||r|r|r|r}
\# Neighbors & Exact & Medium & Fast &  \\
\hline
1 & 4.8  &  119 & 125 &  \\
0.5\% -- 5,242 & 0.2 &  0.2 & 0.2 &  \\
100\% -- $1e6$ & 0.1 & 0.1 & 0.1 & 
\end{tabular}
\caption{Median steps/second as a function of $k$ \& FLANN settings.}
\label{table:puddle}
\end{table}

\subsection{Recommender Task}
\label{sec:exp-rec} 
Experiments were run on 3 different recommender tasks involving 49 elements, 835
elements,  and 13,138 elements.  These tasks' dynamics are quite irregular, with
certain actions being good in many states, and certain states requiring a
specific action rarely used elsewhere. This has the effect of rendering agents with $k=1$ quite poor at this task.  Additionally, although initial exploration methods were purely
uniform random with an epsilon probability, to better simulate the
reality of the running system --- where state transitions are also heavily guided
by user choice --- we restricted our epsilon exploration to a likely subset of good
actions provided to us by the simulator.  This subset is only used to guide
exploration; at each step the policy must still choose amongst the full set of
actions if not exploring. Learning with uniform exploration converges, but in
the larger tasks performance is typically 50\% of that with guided exploration.

\begin{figure}[h]
\includegraphics[width=8cm]{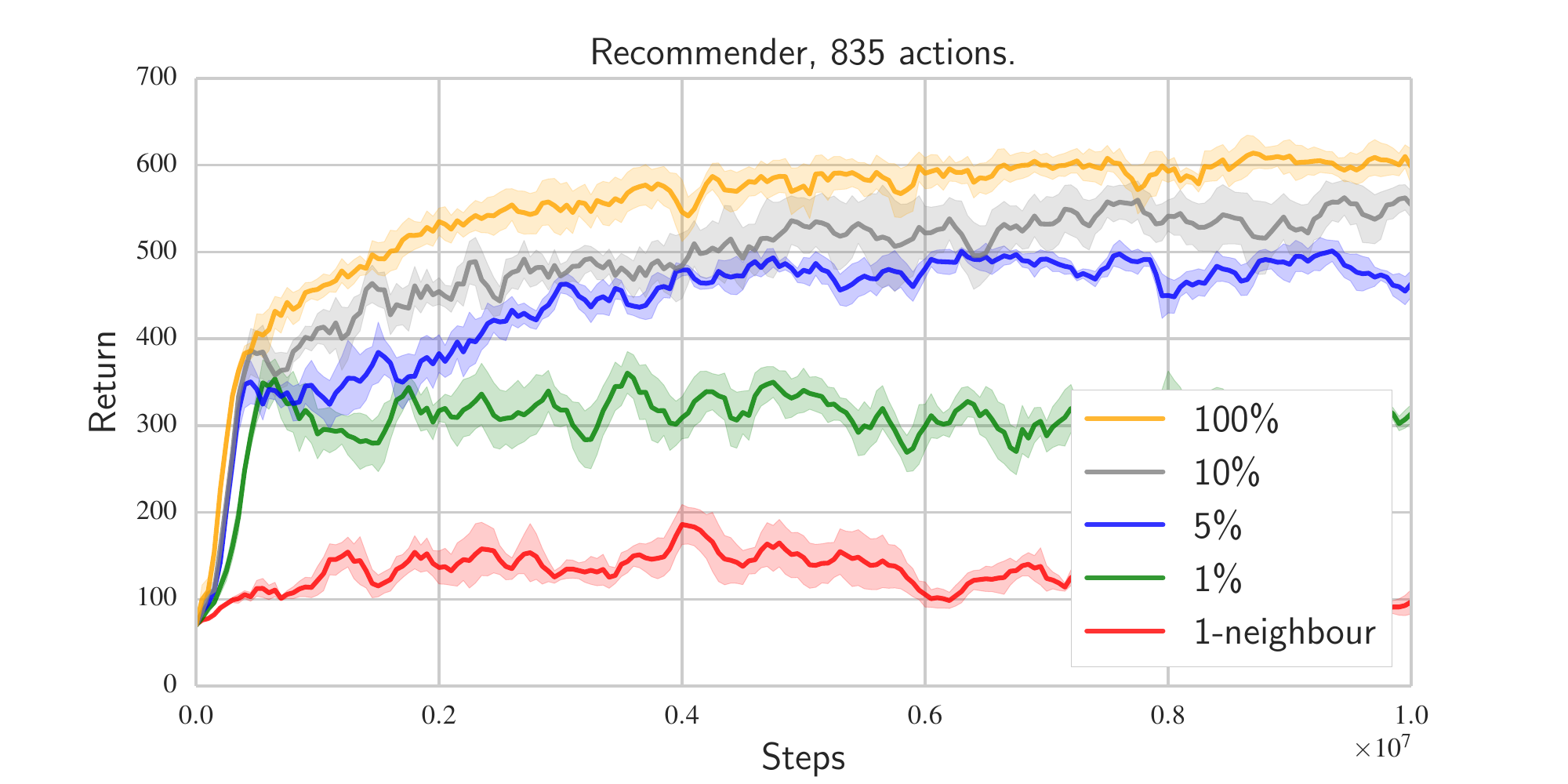}
\caption{Performance on the 835-element recommender task for varying values of $k$, with exact nearest-neighbor lookup.} 
\label{fig:835}
\end{figure}

Figure \ref{fig:835} shows performance on the 835-element task using exact
lookup for varying values of $k$ as a percentage of the total number of actions.
We can see a clear progression of performance as $k$ is increased in this task.  Although not displayed in the plot, these smaller action sizes have much less significant speedups, with $k=|\aset|$ taking only twice as long as $k=83$ (1\%).

\begin{figure}[h]
\includegraphics[width=8cm]{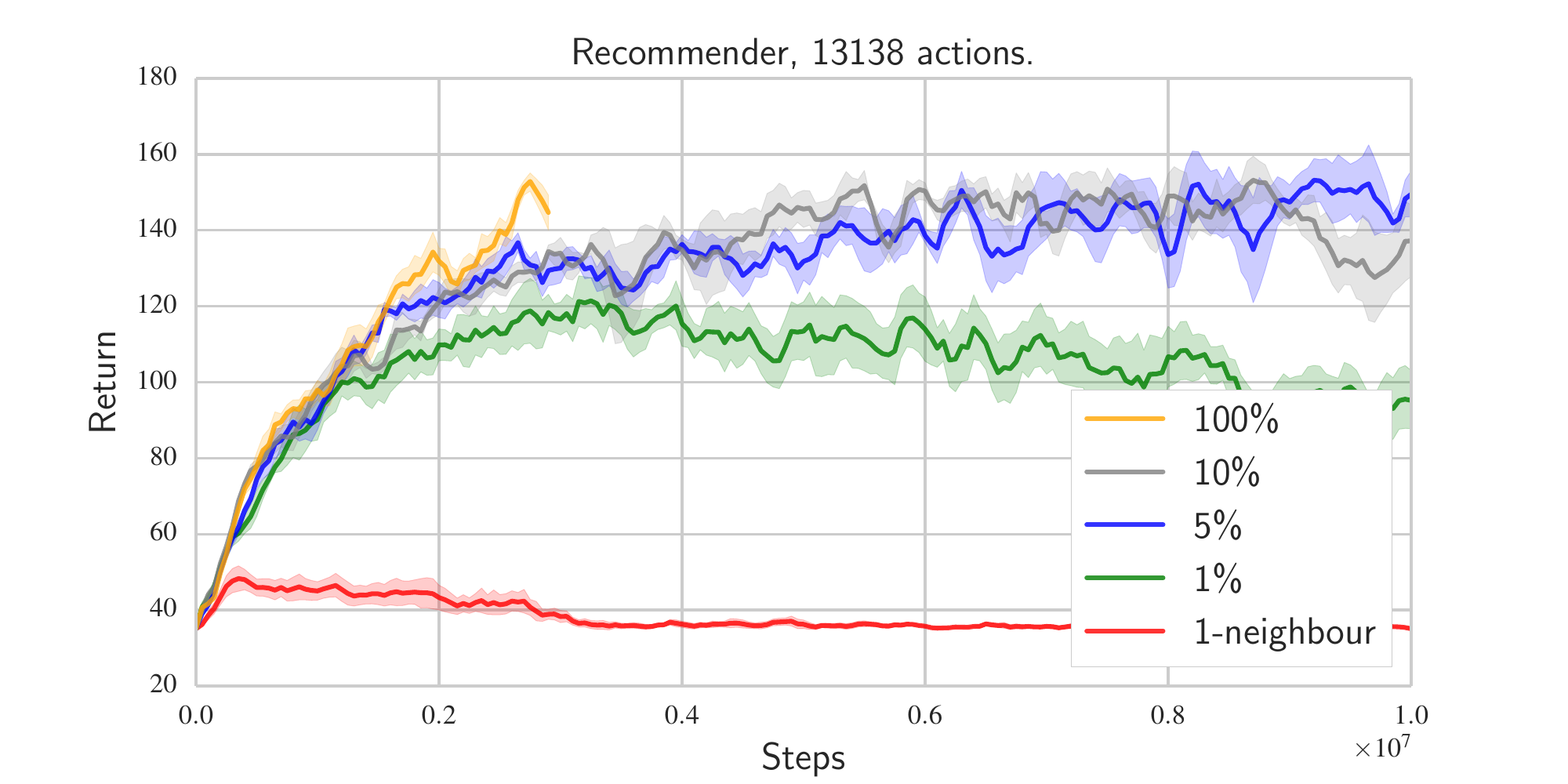}
\caption{Agent performance for various numbers of nearest neighbors on 13k recommender task. Training with $k=1$ failed to learn.}
\label{fig:13k-nn}
\end{figure}

\begin{figure}[h]
\includegraphics[width=8cm]{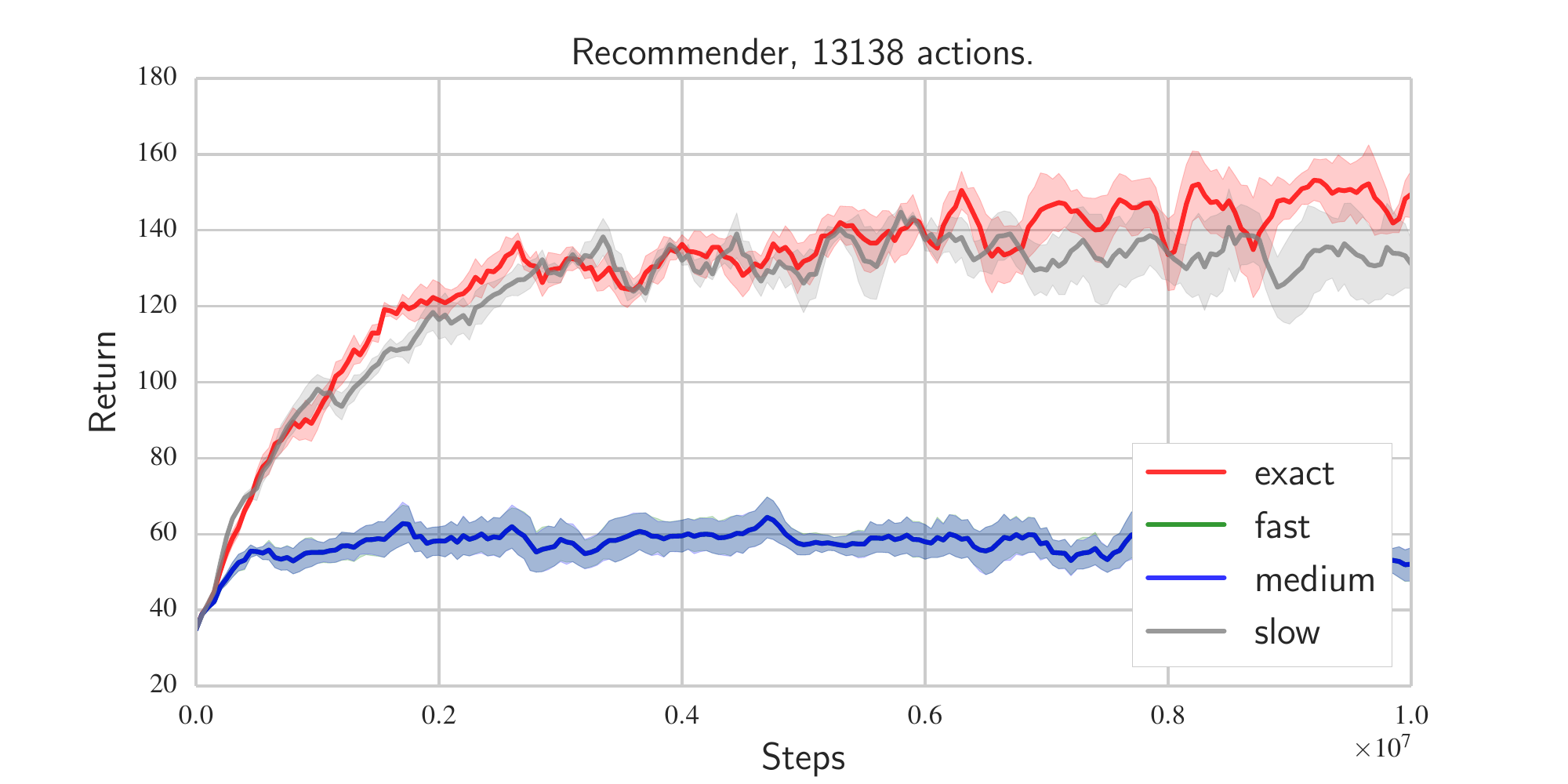}
\caption{Agent performance for various FLANN settings on nearest-neighbor
lookups on the 13k recommender task. In this case, fast and medium FLANN
settings are equivalent. $k=656$ (5\%).} 
\label{fig:13k-flann}
\end{figure}

Results on the $13,138$ element task are visualized in Figures \eqref{fig:13k-nn}
for varying values of $k$, and in Figure \eqref{fig:13k-flann} with varying
FLANN settings.   Figure \eqref{fig:13k-nn} shows performance for exact nearest-
neighbor lookup and varying values of $k$. We note that the agent using all actions
(in yellow) does not train as many steps due to slow training speed.  It is
training approximately 15 times slower in wall-time than the 1\% agent.

Figure \eqref{fig:13k-flann} shows performance for varying FLANN settings on
this task with a fixed $k$ at 5\% of actions. We can quickly see both that
lower-recall settings significantly impact the performance on this task.

\begin{figure}[h]%
    \centering
    \begin{minipage}{0.9\linewidth}
        \begin{figure}[H]
            \includegraphics[width=\linewidth]{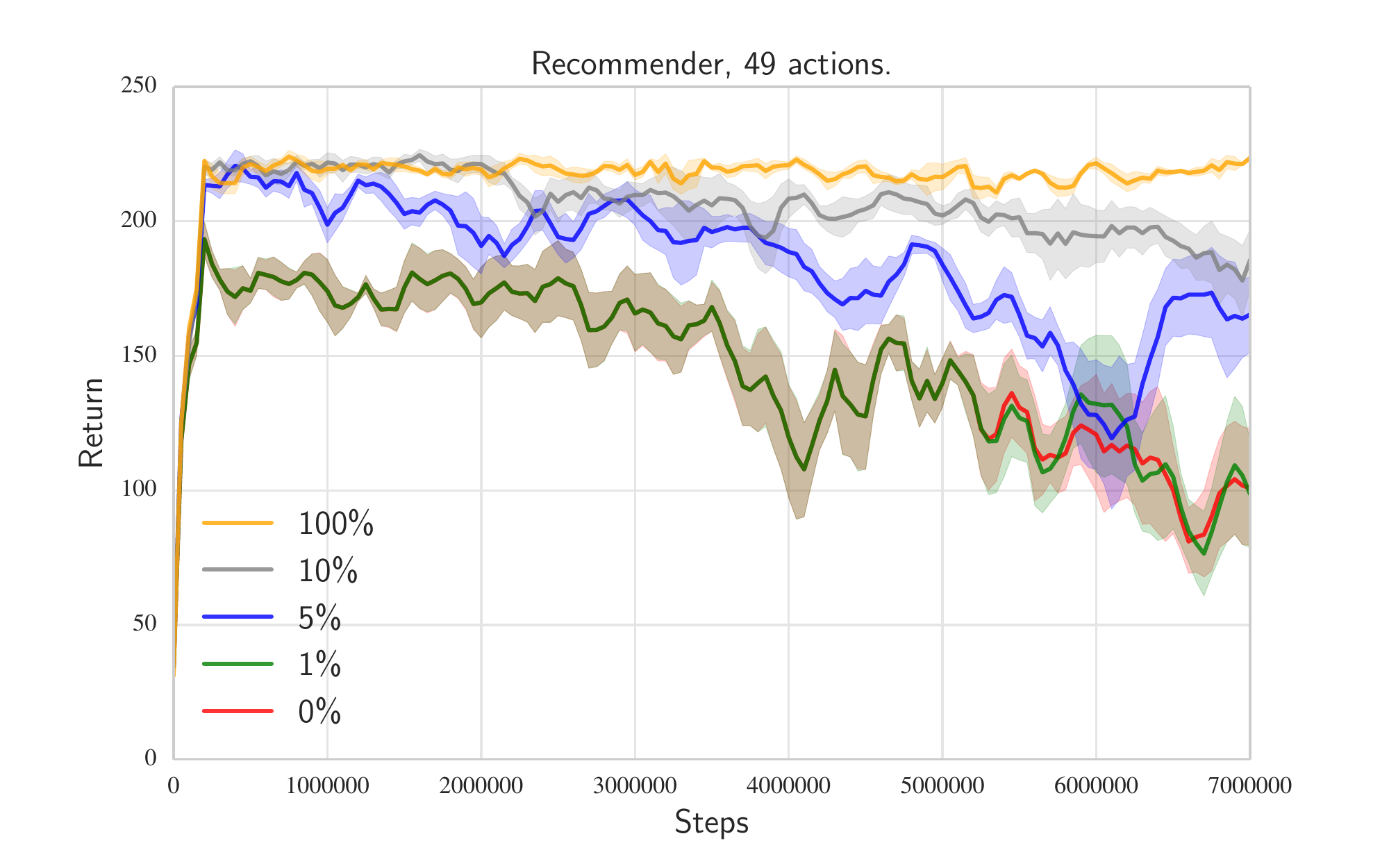}
        \end{figure}
    \end{minipage}\\
    \begin{minipage}{0.9\linewidth}
        \begin{figure}[H]
            \includegraphics[width=\linewidth]{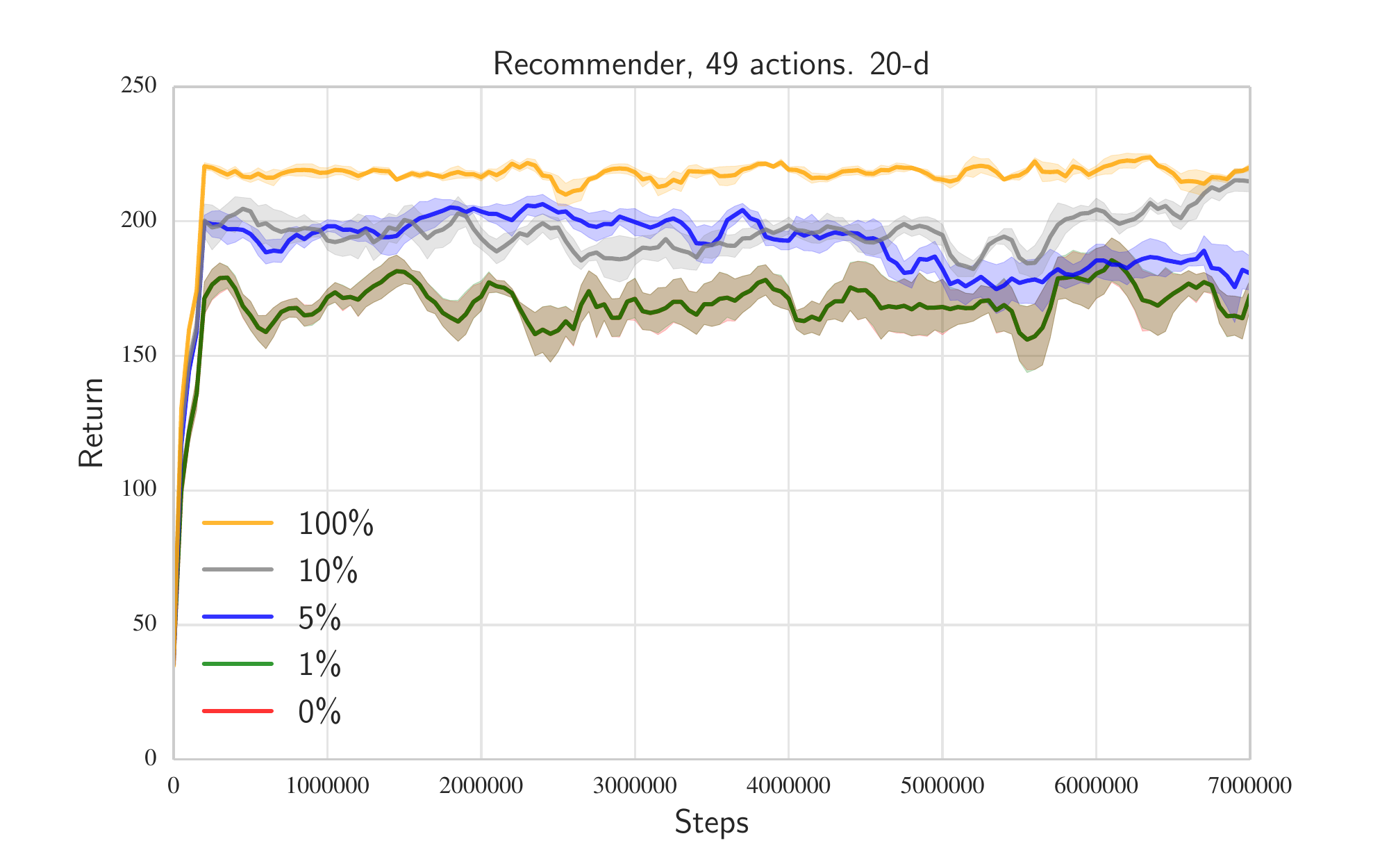}
        \end{figure}
    \end{minipage}
    \caption{Recommender task with 49 actions using 200 dimensional action representation (left) and 20-dimensional action representations (right), for varying values of $k$ and fixed FLANN setting of `slow'.  The figure intends to show general behavior and not detailed values.}
    \label{fig:rec49}
\end{figure}

Results on the 49-element task with both a 200-dimensional and a 20-dimensional
representation are presented in Figure \ref{fig:rec49} using a fixed `slow'
setting of FLANN and varying values of $k$.  We can observe that when using a small
number of actions, a more compact representation of the action space can be
beneficial for stabilizing convergence.

\begin{table}[H]
\centering
\begin{tabular}{r||r|r|r|r}
\# Neighbors & Exact & Slow & Medium & Fast   \\
\hline
1 & 31 &  50 & 69 & 68 \\
1\% --  131  & 23 &  37 & 37 & 37 \\
5\% --  656 & 10 & 13 & 12 &  14 \\
10\% -- 1,313  & 7 & 7.5 & 7.5 & 7 \\
100\% -- 13,138 & 1.5 & 1.6 & 1.5 & 1.4 \\
\end{tabular}
\caption{Median steps/second as a function of $k$ \& FLANN settings on the 13k recommender task.}
\label{table:puddle}
\end{table}
\vspace{-0.5cm}

Results on this series of tasks suggests that our approach can scale to real-world MDPs with large number of actions, but exploration will remain an issue if
the agent needs to learn from scratch.  Fortunately this is generally not the
case, and either a domain-specific system provides a good starting state and action
distribution, or the system's dynamics constrain transitions to a reasonable
subset of actions for a given states.

\section{Conclusion}

In this paper we introduce a new policy architecture able to efficiently learn
and act in large discrete action spaces.  We describe how this architecture can
be trained using DDPG and demonstrate good performance on a series of tasks with
a range from tens to one million discrete actions.

Architectures of this type give the policy the ability to generalize over the set
of actions with sub-linear complexity relative to the number of actions. We
demonstrate how considering only a subset of the full set of actions is
sufficient in many tasks and provides significant speedups.  Additionally, we
demonstrate that an approximate approach to the nearest-neighbor lookup can be
achieved while often impacting performance only slightly.

Future work in this direction would allow the action representations to be
learned during training, thus allowing for actions poorly placed in the
embedding space to be moved to more appropriate parts of the space.  We also
intend to investigate the application of these methods to a wider range of real-world control problems.

\bibliography{paper}

\begin{thebibliography}{16}
\providecommand{\natexlab}[1]{#1}
\providecommand{\url}[1]{\texttt{#1}}
\expandafter\ifx\csname urlstyle\endcsname\relax
  \providecommand{\doi}[1]{doi: #1}\else
  \providecommand{\doi}{doi: \begingroup \urlstyle{rm}\Url}\fi

\bibitem[Dietterich \& Bakiri(1995)Dietterich and
  Bakiri]{dietterich1995solving}
Dietterich, Thomas~G. and Bakiri, Ghulum.
\newblock Solving multiclass learning problems via error-correcting output
  codes.
\newblock \emph{Journal of artificial intelligence research}, pp.\  263--286,
  1995.

\bibitem[Dulac-Arnold et~al.(2012)Dulac-Arnold, Denoyer, Preux, and
  Gallinari]{dulac2012fast}
Dulac-Arnold, Gabriel, Denoyer, Ludovic, Preux, Philippe, and Gallinari,
  Patrick.
\newblock Fast reinforcement learning with large action sets using
  error-correcting output codes for mdp factorization.
\newblock In \emph{Machine Learning and Knowledge Discovery in Databases}, pp.\
   180--194. Springer, 2012.

\bibitem[Hafner \& Riedmiller(2011)Hafner and Riedmiller]{NFQCA}
Hafner, Roland and Riedmiller, Martin.
\newblock Reinforcement learning in feedback control.
\newblock \emph{Machine learning}, 84\penalty0 (1-2):\penalty0 137--169, 2011.

\bibitem[He et~al.(2015)He, Chen, He, Gao, Li, Deng, and Ostendorf]{he2015deep}
He, Ji, Chen, Jianshu, He, Xiaodong, Gao, Jianfeng, Li, Lihong, Deng, Li, and
  Ostendorf, Mari.
\newblock Deep reinforcement learning with an unbounded action space.
\newblock \emph{arXiv preprint arXiv:1511.04636}, 2015.

\bibitem[Lagoudakis \& Parr(2003)Lagoudakis and
  Parr]{lagoudakis2003reinforcement}
Lagoudakis, Michail and Parr, Ronald.
\newblock Reinforcement learning as classification: Leveraging modern
  classifiers.
\newblock In \emph{ICML}, volume~3, pp.\  424--431, 2003.

\bibitem[Lillicrap et~al.(2015)Lillicrap, Hunt, Pritzel, Heess, Erez, Tassa,
  Silver, and Wierstra]{ddpg}
Lillicrap, Timothy~P, Hunt, Jonathan~J, Pritzel, Alexander, Heess, Nicolas,
  Erez, Tom, Tassa, Yuval, Silver, David, and Wierstra, Daan.
\newblock Continuous control with deep reinforcement learning.
\newblock \emph{arXiv preprint arXiv:1509.02971}, 2015.

\bibitem[Lin(1992)]{lin1992self}
Lin, Long-Ji.
\newblock Self-improving reactive agents based on reinforcement learning,
  planning and teaching.
\newblock \emph{Machine learning}, 8\penalty0 (3-4):\penalty0 293--321, 1992.

\bibitem[Mnih et~al.(2015)Mnih, Kavukcuoglu, Silver, Rusu, Veness, Bellemare,
  Graves, Riedmiller, Fidjeland, Ostrovski, et~al.]{mnih15}
Mnih, Volodymyr, Kavukcuoglu, Koray, Silver, David, Rusu, Andrei~A, Veness,
  Joel, Bellemare, Marc~G, Graves, Alex, Riedmiller, Martin, Fidjeland,
  Andreas~K, Ostrovski, Georg, et~al.
\newblock Human-level control through deep reinforcement learning.
\newblock \emph{Nature}, 518\penalty0 (7540):\penalty0 529--533, 2015.

\bibitem[Muja \& Lowe(2014)Muja and Lowe]{flann_pami_2014}
Muja, Marius and Lowe, David~G.
\newblock Scalable nearest neighbor algorithms for high dimensional data.
\newblock \emph{Pattern Analysis and Machine Intelligence, IEEE Transactions
  on}, 36, 2014.

\bibitem[Pazis \& Parr(2011)Pazis and Parr]{pazis2011generalized}
Pazis, Jason and Parr, Ron.
\newblock Generalized value functions for large action sets.
\newblock In \emph{Proceedings of the 28th International Conference on Machine
  Learning (ICML-11)}, pp.\  1185--1192, 2011.

\bibitem[Prokhorov et~al.(1997)Prokhorov, Wunsch,
  et~al.]{prokhorov1997adaptive}
Prokhorov, Danil~V, Wunsch, Donald~C, et~al.
\newblock Adaptive critic designs.
\newblock \emph{Neural Networks, IEEE Transactions on}, 8\penalty0
  (5):\penalty0 997--1007, 1997.

\bibitem[Silver et~al.(2014)Silver, Lever, Heess, Degris, Wierstra, and
  Riedmiller]{DPG}
Silver, David, Lever, Guy, Heess, Nicolas, Degris, Thomas, Wierstra, Daan, and
  Riedmiller, Martin.
\newblock Deterministic policy gradient algorithms.
\newblock In \emph{Proceedings of The 31st International Conference on Machine
  Learning}, pp.\  387--395, 2014.

\bibitem[Sunehag et~al.(2015)Sunehag, Evans, Dulac-Arnold, Zwols, Visentin, and
  Coppin]{sunehag2015deep}
Sunehag, Peter, Evans, Richard, Dulac-Arnold, Gabriel, Zwols, Yori, Visentin,
  Daniel, and Coppin, Ben.
\newblock Deep reinforcement learning with attention for slate markov decision
  processes with high-dimensional states and actions.
\newblock \emph{arXiv preprint arXiv:1512.01124}, 2015.

\bibitem[Sutton \& Barto(1998)Sutton and Barto]{sutton1998rl}
Sutton, Richard~S and Barto, Andrew~G.
\newblock \emph{Reinforcement learning: An introduction}, volume~1.
\newblock MIT press Cambridge, 1998.

\bibitem[Todorov et~al.(2012)Todorov, Erez, and Tassa]{todorov2012mujoco}
Todorov, Emanuel, Erez, Tom, and Tassa, Yuval.
\newblock Mujoco: A physics engine for model-based control.
\newblock In \emph{Intelligent Robots and Systems (IROS), 2012 IEEE/RSJ
  International Conference on}, pp.\  5026--5033. IEEE, 2012.

\bibitem[Van~Hasselt et~al.(2009)Van~Hasselt, Wiering, et~al.]{van2009using}
Van~Hasselt, Hado, Wiering, Marco, et~al.
\newblock Using continuous action spaces to solve discrete problems.
\newblock In \emph{Neural Networks, 2009. IJCNN 2009. International Joint
  Conference on}, pp.\  1149--1156. IEEE, 2009.

\end{thebibliography}
\bibliographystyle{icml2015}

\begin{appendices}
\section{Detailed Wolpertinger Algorithm}
\label{sec:full_algorithm}
Algorithm \ref{alg:ddpg} describes the full DDPG algorithm with the notation used in our paper, as well as the distinctions between actions from $\aset$ and prototype actions.
\begin{algorithm}[h]
  \caption{Wolpertinger Training with DDPG\label{algo:ddpg}}
  \label{alg:ddpg}
  \begin{algorithmic}[1]
    \STATE Randomly initialize critic network $Q_{\theta^Q}$ and actor
    $f_{\theta^\pi}$ with weights $\theta^Q$ and $\theta^\pi$.
    \STATE Initialize target network $Q_{\theta^Q}$ and $f_{\theta^\pi}$ with weights ${\theta^Q}'
    \leftarrow \theta^Q, {\theta^\pi}' \leftarrow \theta^\pi$

    \STATE Initialize $g$'s dictionary of actions with elements of $\aset$
    \STATE Initialize replay buffer $B$
    \FOR{episode = 1, M}
      \STATE Initialize a random process $\mathcal{N}$ for action
      exploration
      \STATE Receive initial observation state $\state_1$
      \FOR{t = 1, T}
        \STATE Select action $\action_t = \pi_\theta(\state_t)$
        according to the current policy and exploration method
        \STATE Execute action $\action_t$ and observe
        reward $r_t$ and new state $\state_{t+1}$
        \STATE Store transition $(\state_t, \action_t, r_t, \state_{t+1})$ in $B$
        \STATE Sample a random minibatch of $N$ transitions
               $(\state_i, \action_i, r_i, \state_{i + 1})$ from $B$
        \STATE Set $y_i = r_i + \gamma \cdot Q_{\theta^{Q'}}(\state_{i + 1},
        \pi_{\theta'}(\state_{i+1}))$
        \STATE Update the critic by minimizing the loss:\\
               $L(\theta^Q) = \frac{1}{N} \sum_i [y_i -
               Q_{\theta^Q}(\state_i, \action_i)]^2$
        \STATE Update the actor using the sampled gradient:
        \begin{align*}
            &\nabla_{\theta^{\pi}} f_{\theta^\pi}|_{\state_i} \approx \\
            &\frac{1}{N} \sum_i
               \nabla_{\action} Q_{\theta^Q}(\state, \hat{\action})|_{\hat{\action} = f_{\theta^\pi}(\state_i)} \cdot
               \nabla_{\theta^{\pi}} f_{\theta^{\pi}}(\state)|_{\state_i}
         \end{align*}
        \STATE Update the target networks:
          \begin{align*}
            {\theta^Q}' &\leftarrow \tau {\theta^Q} + (1 - \tau) {\theta^Q}'\\
            {\theta^\pi}' &\leftarrow \tau {\theta^\pi} +
                (1 - \tau) {\theta^\pi}'
          \end{align*}
        \ENDFOR
    \ENDFOR
  \end{algorithmic}
\end{algorithm}

The critic is trained from samples stored in a replay buffer.  These samples are generated on lines 9 and 10 of Algorithm
\ref{alg:ddpg}.  The action $\action_t$ is sampled from the full Wolpertinger
policy $\pi_\theta$ on line 9.  This action is then applied on the environment
on line 10 and the resulting reward and subsequent state are stored along with
the applied action in the replay buffer on line 11.

On line 12, a random transition is sampled from the replay buffer, and line 13
performs Q-learning by applying a Bellman backup on
$Q_{\theta^Q}$, using the target network's weights for the target Q.  Note the target action is generated by the full policy $\pi_\theta$ and not simply $f_{\theta^\pi}$.

The actor is then trained on line 15 by following the policy gradient:

\begin{align*}
\nabla_{\theta}f_{\theta^\pi} &\approx \E_{f'}\left[\nabla_{\theta^\pi} 
Q_{\theta^Q}(\state, \hat{\action})|_{\hat{\action} = f_\theta(\state)} \right]\\
&  = \E_{f'}\left[\nabla_{\hat{\action}} Q_{\theta^Q}(\state, f_\theta(\state)) \cdot \nabla_{\theta^\pi} f_{\theta^\pi}(\state)|\right].
\end{align*}

Actions stored in the replay buffer are generated by $\pi_{\theta^\pi}$, but the
policy gradient $\nabla_{\hat{\action}} Q_{\theta^Q}(\state, \hat{\action})$ is taken at
$\hat{\action} = f_{\theta^\pi}(\state)$.  This allows the learning algorithm to
leverage otherwise ignored information of which action was actually executed for
training the critic, while taking the policy gradient at the actual output of
$f_{\theta^\pi}$.

\section{Proof of Lemma 1}
\begin{proof}
Without loss of generality we can assume $Q(s,a) = \frac{1}{2}$, $b=\frac{1}{2}$ and replace $c$ with $c' = \frac{c}{2b}$, resulting in an affine transformation of the original setting.  We undo this transformation at the end of this proof to obtain the general result.

There is a $p$ probability that an action is `bad' and has value $-c'$.  If it is not bad, the distribution of the value of the action is uniform in $[Q(s,a)-b, Q'(s,a)+b] = [0,1]$.  This implies that the cumulative distribution function (CDF) for the value of an action $i \in \{1, \ldots k \}$ is
\[
F(x;s,i)
=
\left\{\begin{array}{ll}
0 & \text{for $x < -c$}\\
p &  \text{for $x \in [-c, 0)$} \\
p + (1 - p)x &  \text{for $x = [0,1]$} \\
1 & \text{for $x > 1$} \,.
\end{array}\right.
\]
If we select $k$ such actions, the CDF of the maximum of these actions equals the product of the individual CDFs, because the probability that the maximum value is smaller that some given $x$ is equal to the probability that all of the values is smaller than $x$, so that the cumulative distribution function for 
\def\Fmax{F_{\max}(x;s,a)}
\begin{align*}
\Fmax
& = P\left(\max_{i \in \{1, \ldots k\}} Q(s,i) \le x\right) \\
& = \prod_{i\in\{1, \ldots, k\}} P\left(Q(s,i) \le x\right) \\
& = \prod_{i\in\{1, \ldots, k\}} F(x;s,i) \\
& = F(x;s,1)^k \,,
\end{align*}
where the last step is due to the assumption that the distribution is equal for all $k$ closest actions (it is straightforward to extend this result by making other assumptions, e.g., about how the distribution depends on distance to the selected action).
The CDF of the maximum is therefore given by
\[
\Fmax
=
\left\{\begin{array}{ll}
0 & \text{for $x < -c'$}\\
p^k & \text{for $x \in [-c', 0)$}\\
\left(p + (1 - p)x\right)^k &  \text{for $x \in [0,1]$} \\
1 & \text{for $x > 1$} \,.
\end{array}\right.
\]
Now we can determine the desired expected value as
\begin{align*}
& \mathbb{E}[\max_{i \in \{1, \ldots, k\}} Q(s,i)] \\
& = \int_{-\infty}^\infty x  \, \text{d} \Fmax \\
& = p^k \left(\frac{1}{2}-c'\right) + \int_{0}^1 x \, \text{d} \Fmax \\
& = p^k \left(\frac{1}{2}-c'\right) + \left[ x \Fmax \right]_0^1 -\int_{0}^1 \Fmax \, \text{d} x \\
& = p^k \left(\frac{1}{2}-c'\right) + 1 - \int_{0}^1 \left(p + (1 - p)x\right)^k \, \text{d} x \\
& = p^k \left(\frac{1}{2}-c'\right) + 1 - \left[ \frac{1}{1 - p}\frac{1}{k+1}\left(p + (1 - p)x\right)^{k+1} \right]_0^1 \\
& = p^k \left(\frac{1}{2}-c'\right) + 1 - \left( \frac{1}{1 - p}\frac{1}{k+1} - \frac{1}{1 - p}\frac{1}{k+1}p^{k+1} \right) \\
& = 1 + p^k \left(\frac{1}{2} - c'\right) - \frac{1}{k+1} \frac{1 - p^{k+1}}{ 1 - p } \,,
\end{align*}
where we have used $\int_0^1 x \,\text{d} \mu(x) = \int_0^1 1 - \mu(x) \,\text{d}x$, which can be proved by integration by parts.
We can scale back to the arbitrary original scale by subtracting $1/2$, multiplying by $2b$ and then adding $Q(s,a)$ back in, yielding
\begin{align*}
&\mathbb{E}\left[\max_{i \in \{1, \ldots, k\}} Q(s,i)\right]\\
& = Q(s,a) + 2 b \left( 1 + p^k \left( \frac{1}{2} - c'\right) - \frac{1}{k+1} \frac{1 - p^{k+1}}{ 1 - p } - \frac{1}{2} \right) \\
& = Q(s,a) + b + p^k b - p^k c - \frac{2 b}{k+1} \frac{1 - p^{k+1}}{ 1 - p } \\
& = Q(s,a) + b  - p^k c - b \left(\frac{2}{k+1} \frac{1 - p^{k+1}}{ 1 - p } - p^k\right)
\qedhere
\end{align*}
\end{proof}
\end{appendices}

\end{document}